\documentclass[12pt]{article}

\usepackage[english]{babel}
\usepackage{csquotes}
\usepackage[
  style=apa,
  backend=biber,
  sorting=nyt,
  maxcitenames=2,
  maxbibnames=99,
  uniquelist=false
]{biblatex}
\DeclareLanguageMapping{english}{english-apa}
\usepackage[hidelinks]{hyperref}
\usepackage{setspace}
\usepackage{microtype}
\title{Evaluating the Use of Large Language Models as Synthetic Social Agents in Social Science Research}
\author{Emma Rose Madden\thanks{University of Oxford, Department of Politics and International Relations. Email: emma.madden@politics.ox.ac.uk}}
\date{September 2025}

\begin{document}
\maketitle

\doublespacing
\setlength{\parindent}{14pt}
\setlength{\parskip}{0pt}

\begin{abstract}
\begin{singlespace}
Large Language Models (LLMs) are being increasingly used as synthetic agents in social science, in applications ranging from augmenting survey responses to powering multi-agent simulations. This paper outlines cautions that should be taken when interpreting LLM outputs and proposes a pragmatic reframing for the social sciences in which LLMs are used as high-capacity pattern matchers for quasi-predictive interpolation under explicit scope conditions and not as substitutes for probabilistic inference. Practical guardrails such as independent draws, preregistered human baselines, reliability-aware validation, and subgroup calibration, are introduced so that researchers may engage in useful prototyping and forecasting while avoiding category errors.

\end{singlespace}
\end{abstract}

\section{Introduction}

The advent of Large Language Models (LLMs) and their potential to act as \textit{homo silicus} has sparked considerable excitement across various social science disciplines \parencite{horton2023}. Researchers have begun leveraging these models to simulate human populations, behaviors, and social interactions \parencite{ArgyleBusbyFuldaGublerRyttingWingate2023, binz2023turninglargelanguagemodels, hewitt2024, park2023generativeagentsinteractivesimulacra, park2024generativeagentsimulations1000}.The growing momentum in this field suggests we may be on the cusp of a paradigm shift in social science methodology, but it also raises urgent questions about the soundness of this approach.

A critical epistemological danger may lie in how we interpret LLM outputs. Querying LLMs as if they operate like humans, drawing from a stable, real-world distribution, can lead to false conclusions. LLMs do generate text probabilistically by choosing each word based on learned probability distributions, but they are not probabilistic inference engines in the rigorous Bayesian sense. In orthodox predictive methods, uncertainty is quantified and propagated through models, as in Bayesian frameworks where priors are updated into posteriors given new data \parencite{gelman2007, pearl2009}. An LLM, by contrast, does not maintain or update a posterior belief distribution when given new prompts; it has no explicit model of uncertainty or variance in the phenomena it is simulating. Its “probabilistic” output is actually the result of pre-trained static weights and sampling algorithms such as temperature or top-\textit{k}/\textit{p}, and not an updating mechanism that incorporates new evidence. When an LLM is instructed to act as, for example, a 30-year-old Midwestern farmer or a college-educated urban voter, it is not drawing from an actual demographic population distribution. Rather, the LLM is acting as a complex pattern-matching system that imitates what such a person might say based on traces in the LLM’s training data. If LLMs do not truly reason over distributions of people’s beliefs in the way social scientists do, then what exactly is being inferred when their outputs are treated as if they were samples from a population?

This paper addresses two central questions stemming from this discrepancy: What are the statistical limitations on the value of using LLMs as synthetic agents in social research, and what does the future of LLM agents look like in the field?

\section{LLMs as Predictive Synthetic Agents}

Recent work has treated LLM-as-respondent as a low-cost proxy for real human sampling in surveys and lab studies. \textcite{ArgyleBusbyFuldaGublerRyttingWingate2023} simulate demographic subpopulations by conditioning GPT-3 on rich backstories and compare the resulting LLM-generated silicon samples to human survey distributions. \textcite{aher2023usinglargelanguagemodels} propose experiments in which models generate many participant-like responses to classic findings in economics and psychology, documenting both replications and systematic distortions. \textcite{park2024generativeagentsimulations1000} introduce a survey automation framework that uses LLMs to design, field, and analyze surveys. They validate the approach against human data, showing that LLM-generated responses closely replicate patterns observed in the U.S. General Social Survey (GSS).

Other studies evaluate LLMs as predictive tools for social-behavioral results. \textcite{hewitt2024} show that GPT-4 can anticipate the direction and relative strength of treatment effects across a wide range of social science experiments, including unpublished studies, and in many cases does so more reliably than human experts. \textcite{chen2025predictingfieldexperimentslarge} similarly find that LLMs can predict the outcomes of economic field experiments, but they also highlight important blind spots, particularly in areas involving sensitive social categories such as gender, ethnicity, or norms. \textcite{yeykelis2025usinglargelanguagemodels} generate AI-powered personas to simulate media-effects experiments, reproducing a majority of main findings and demonstrating the potential and limits of LLM-based experimental replication.

Additionally, agent-based lines of work embed LLMs in multi-agent simulations to study interactional dynamics, shifting from LLM-as-respondent to LLM-as-agent. \textcite{park2023generativeagentsinteractivesimulacra} show that memory-, planning-, and reflection-augmented agents produce believable micro- and meso-level social behavior in sandbox environments, which some argue could serve as testbeds for hypothesis generation or intervention prototyping. At the same time, ambitious claims that foundation models can “predict and simulate human behavior in any experiment expressible in language” \parencite{binz2025nature} have already attracted pointed methodological critiques, underlining active debate over what exactly is being predicted.

\section{The Illusion of Probabilism in LLMs}

LLMs generate responses by auto-regressively predicting tokens based on learned distributional patterns from massive datasets conditioned by user-input prompts \parencite{brown2020languagemodelsfewshotlearners}. LLMs employ various sampling techniques, such as temperature scaling, top-\textit{k}, and top-\textit{p}, that introduce variability and allow for flexibility in model output \parencite{bommasani2022opportunitiesrisksfoundationmodels,vaswani2017attention}. Toggling these metrics results in variation in the output text, even while prompting is held consistent. The resulting text reflects conditional plausibility under the model, not draws from a well-defined posterior over a human population.

In certain controlled settings, LLMs can learn to mimic the optimal Bayesian predictor for a given task \parencite{panwar2024incontextlearningbayesianprism}. However, in this circumstance there also exists a “forgetting phenomenon” where a model first learns to generalize like a Bayesian but then reverts to memorizing its pre-training data as training continues. This suggests that while transformers can learn to be Bayesian predictors, that ability is contingent on the training environment and not an inherent property of the model.

A core tenet of statistical modeling with independent and identically distributed (i.i.d.) or, more generally, exchangeable data is that the order in which observations are presented should not affect inference. Social science data, such as survey responses or experimental results, are typically treated as exchangeable. However, repeated queries of LLMs can fail this invariance test.

\textcite{falck2024incontextlearninglargelanguage} trace this failure to a violation of a key condition for Bayesian learning that they term the martingale property of the predictive distribution (equivalently, forecast invariance/CID). For a sequence of random variables $Z_i$, the martingale property states that the expectation of the next value, given all previous values, is equal to the expectation of a value much further in the future, given the same history. For a model $p_M$, any $n$, any $k\ge 1$, and all $z$,
\begin{equation}
  p_M\!\left(Z_{n+1}=z \,\middle|\, Z_{1:n}\right)
  \;=\;
  p_M\!\left(Z_{n+k}=z \,\middle|\, Z_{1:n}\right).
  \label{eq:forecast-invariance}
\end{equation}

This property ensures that a model’s predictive distribution is invariant to imputations of missing samples from its own distribution on average. When this property is violated, the model’s predictions become incoherent and ambiguous. If an agent’s probabilistic beliefs are coherent, then they can be represented as Bayesian probabilities. In other words, a model or agent is Bayesian if and only if its probability assignments are coherent and exchangeable/conditionally identically distributed (CID) \parencite{definetti1974}).  In theory, an LLM could be Bayesian if it could be shown to produce coherent predictions, but as of yet there exists no consistent evidence of this happening in practice. An LLM might produce different marginal predictions for the 100th participant in a simulated trial depending on whether it first imputed the 51st or 99th participant. Such order dependence undermines the credibility of LLM-generated data in scenarios where the order of observation is known to be irrelevant.

The violation of the martingale property leads to a phenomenon that \textcite{falck2024incontextlearninglargelanguage} call “introspective hallucination”. This occurs when a model systematically changes its own predictions on average by querying itself and generating new data points to append to its context. This is a direct consequence of the model’s predictions drifting rather than converging as a true Bayesian posterior would. The study presented specific diagnostics to test for this. For example, the test statistic $T_1,g$ measures the average difference between predictions at different future time steps. For a system satisfying the martingale property, this statistic should be centered around zero. Their experiments showed a consistent, non-zero drift for models like GPT-4, Llama-2, and Mistral-7B, especially on longer data sequences, giving evidence of this non-Bayesian behavior \parencite{falck2024incontextlearninglargelanguage}.

This finding is critical for social scientists hoping to use synthetic agents in their research. If an LLM is used to simulate a population by generating one agent at a time and adding them to the context, its predictive distribution should update as more agents are introduced, so later draws can legitimately differ from earlier ones. What must not happen, however, is for predictions to depend on the arbitrary order in which those agents were generated. This order-insensitivity is a necessary condition for unambiguous prediction under exchangeable data \parencite{falck2024incontextlearninglargelanguage}. In a coherent model, the sequence of imputations should be CID: before observing any new sample, the prediction for the 200th agent should be the same as for the 20th, and once a new sample is imputed, the distribution is consistently updated given the expanded history. This requirement is a natural generalization of de Finetti’s exchangeability \parencite{definetti1974} and has been formalized in the probability literature on predictive characterizations of Bayesian models \parencite{Berti_2004,fortini2012}. By contrast, if an LLM’s predictions vary with the order of imputation rather than only with the information contained in past samples, it undermines the credibility of the simulated data and risks introducing serious biases into research findings.

\section{Why LLMs Seem To “Predict”}

LLMs’ increasing realism and improved predictive capabilities emerge largely due to scale and architecture. At their core, LLMs are unparalleled pattern matchers. They are trained on massive datasets nearing the scale of the internet, which represents a significant portion of recorded human language and knowledge. Both apparent and latent in this training data are the social patterns, heuristics, biases, attitudes, and behavioral tendencies that form the basis of social science research. When an LLM is prompted to predict a behavior or attitude, it is not performing inference in a statistical sense. Instead, it is pulling from learned patterns of natural language. The extent to which natural language acts as an accurate representation of lived reality (ala \textcite{wittgenstein1953}), is still a topic of philosophical debate.

However, LLMs have demonstrated the capability to “predict” human behavior in experiments \parencite{aher2023usinglargelanguagemodels, ArgyleBusbyFuldaGublerRyttingWingate2023,park2023generativeagentsinteractivesimulacra,park2024generativeagentsimulations1000} under various prompting, steering, and tuning strategies. In each case, the LLM’s success hinges on its ability to interpolate from the massive distribution of data it has already seen. In these cases, and potentially in other social phenomena that are widely discussed and well documented, pattern matching may just be good enough to produce outputs that mimic reality.

LLM performance is subject to scaling laws, which show that as you predictably increase the number of model parameters and the size of the training dataset, performance on various benchmarks improves \parencite{kaplan2020scalinglawsneurallanguage}. Scale gives rise to emergent abilities—capabilities not present in smaller models but that appear spontaneously in larger ones \parencite{wei2022emergentabilitieslargelanguage}. Larger models should, therefore, allow for more accurate simulation of human dialogue, which would in turn lead to increased “predictive” capability. Amongst these emergent phenomena even exists probabilistic mimicry. This has led to competing explanations for how such mimicry works. A leading theoretical explanation for this is implicit Bayesian inference \parencite{xie2022explanationincontextlearningimplicit}, which holds that the model learns to infer a high-level “latent concept” from the examples in a prompt. The model’s behaviors can then be described as a Bayesian update when it conditions its output on this inferred concept. An alternative, more mechanistic explanation comes by way of \textcite{akyurek2024incontextlanguagelearningarchitectures}, who emphasize that transformer performance in in-context learning derives from their capability to mimic probabilistic methods implicitly rather than explicitly performing Bayesian inference.

On top of the base models, most researchers using LLMs for social sciences are using additional methods to fine-tune or steer model output. Off-the-shelf LLMs are known to produce inherently biased outputs based on biased training data \parencite{santurkar2023opinionslanguagemodelsreflect}. Methods like fine-tuning and reinforcement learning from human feedback (RLHF) refine model outputs by optimizing them to align with human behavior in specific contexts, such as on cognitive tasks or in behavioral games \parencite{binz2025nature}. This makes the models better conversationalists and more adept at following instructions, but it is a process of behavioral alignment and not statistical correction. This distinction raises a fundamental question for social science: when we tune a model to be better at a specific human task, like replicating survey results, are we creating a more generalizable model of human behavior, or are we simply creating a more specialized mimic?

This question forces a confrontation with the classic academic tension between explanation and prediction. Fine-tuning an LLM on a specific dataset, such as behavioral economic games, may indeed make it highly proficient at predicting outcomes in that narrow context. However, this optimization does not guarantee the model has developed a deeper, more generalizable understanding of human motivations like fairness, reciprocity, or strategic thinking. Instead, the model may be overfitting to the statistical patterns of the training data, much like a regression model with too many parameters. This creates a brittle agent that excels at its trained task but might fail spectacularly when presented with a novel scenario. Here, social scientists might question this line of research’s ability to provide insight into real and complex human behavior at a more general level of analysis.

\textcite{falck2024incontextlearninglargelanguage} provides strong empirical evidence for this trade-off. They found that LLMs that had undergone extensive instruction tuning and alignment (like GPT-3.5 and GPT-4) actually generally performed worse on their tests of statistical coherence than older, less-tuned foundational models. This suggests that the very process of making a model a better conversationalist and instruction-follower may degrade its adherence to fundamental Bayesian principles. This raises a provocative methodological question: is the act of fine-tuning an LLM to replicate a dataset a form of social science? Or is it an exercise in creating a quasi-predictive lookup system? Traditional social science aims to build and test theories that can explain behavior across different contexts. If our synthetic agents are tuned to the point where they can only reproduce the data they were trained on, we risk engaging in a circular exercise that discovers no new knowledge and offers no generalizable theory.

In this lies a warning for the field— a single, one-size-fits-all \textit{homo silicus} is likely a mirage. An agent fine-tuned to be a perfect respondent for political surveys may be a poor model for studying cognitive biases, and both may be unsuited for exploring long-term social dynamics. The future of synthetic agent-based modeling will require not a universal tool, but a multitude of micro-level, domain-specific approaches where researchers are transparent about the optimizations they have made and deeply critical of the generalizability they may have sacrificed.

\section{Making Robust Synthetic Agents}

With the aforementioned warnings in mind, some practical guidelines should be set forth on how to deploy LLM-based synthetic agents with scientific rigor. First, and as previously emphasized, researchers should treat LLM-backed agents as high-capacity predictors of human responses under specified conditioning and not as sources of calibrated probabilistic belief. A good example of this is how \textcite{ArgyleBusbyFuldaGublerRyttingWingate2023} operationalize “algorithmic fidelity”—defined as the extent to which a conditioned model reproduces the conditional association structure among attitudes, demographics, and behaviors observed in matched human groups rather than just matching marginals or surface text—and validate those associations against real surveys by generating one silicon subject per backstory. This means that queries from their agents act more as scoped interpolation as opposed to as posterior claims. Calibration and subgroup validity must be actively measured and reported and not just assumed. Algorithmic fidelity is conditional and demographically structured, so accuracy can vary across groups \parencite{ArgyleBusbyFuldaGublerRyttingWingate2023}. Researchers should report the dispersion of synthetic responses and their accuracy relative to human baselines for each salient subgroup rather than only overall averages.

Empirically, \textcite{aher2023usinglargelanguagemodels} show that their Turing experiments reproduce familiar effects, but repeated samples exhibit “hyper-accuracy distortion,” or more specifically, hyper-accuracy variance collapse. The model’s answers cluster too tightly and are consistently closer to the keyed truth than real people would be, so the variability that would be expected from a human population is missing. \textcite{Bisbee_Clinton_Dorff_Kenkel_Larson_2024} also observed unnaturally low variance and misaligned regression coefficients in LLM-generated survey data. Departures from human-like variance such as this should be flagged in future research. Since the spread is artificially small, any posterior-style inference from repeated draws is misleadingly precise. Whenever an LLM’s silicon sample is too confident or too homogeneous compared to humans, researchers should down-weight that synthetic evidence or refrain from treating it as a reliable proxy in statistical analyses.

Any inference from LLM-simulated data should acknowledge distribution shift and that these are model outputs, not draws from a human population. Because models can deviate systematically from human opinions, growing a synthetic sample from itself can produce intervals that are too narrow. A better approach is to construct confidence sets that reflect mismatch and to adaptively cap the effective simulation size. \textcite{huang2025uncertaintyquantificationllmbasedsurvey} provide a framework that selects a data-driven size \(\hat{k}\) and shows that taking \textit{k} very large can hurt coverage, while the selected \(\hat{k}\) maintains near-nominal coverage. This shows that alignment with human data should determine interval tightness. These types of techniques are model agnostic and use synthetic data to augment human data instead of replacing it. Augmentation offers a promising path forward in its own right. For example, synthetic agents have the potential to be used to pretest instruments, surface subgroup calibration issues, identify failure modes before fielding, or guide sample allocation and sensitivity analysis. In this, scarce human measurements can be collected where they are most informative. This could raise power and lower cost at the research design stage and still preserve human data as the standard for inference.

A practical way to begin to tackle the discussed statistical discrepancies is to construct agents from individual-level human data rather than from generic personas. For example, \textcite{park2024generativeagentsimulations1000} build interview-anchored agents and show strong agreement with individuals’ GSS responses and several preregistered replications, approaching human test–retest reliability; ablations conducted indicate the gains come from anchoring rather than a sort of stylistic mimicry. These choices clarify scope and provide natural validation targets, yet \textcite{park2024generativeagentsimulations1000} also report heterogeneity by anchoring method and task, which cautions against generalization beyond the elicited domain. They further find that interview-based agents reduce bias gaps across political, racial, and gender groups compared to simpler demographic-only prompts, indicating improved calibration, though certain subgroup predictions can still deviate from human variance. Algorithmic fidelity is not all-or-nothing; it should be quantified per subgroup, with any over- or under-dispersion and miscalibration relative to human benchmarks transparently reported.

Similar logic can carry from person anchoring to task anchoring at scale. \textcite{binz2025nature}’s CENTAUR fine-tunes on large, trial-by-trial behavioral datasets with explicit held-out structures and reports out-of-domain generalization. This transparency in task grounding and rigorous out-of-domain testing exemplifies a predictive framing for LLM agents. At the same time, even a model as powerful as CENTAUR remains constrained by its training domain, so strong performance on experimental tasks does not license broad posterior-style claims about human nature writ large. Fine-tuned success on specific task families should be interpreted narrowly as a high-capacity predictor for those contexts rather than a universal model for \textit{homo silicus}. As compute and tooling improve, individual-level fine tuning may become feasible for consenting participants, but the real bottleneck is longitudinal, high-quality personal data. Even then, context and conditioning are key, because individual tuning mostly reduces between-person noise and shifts the uncertainty to within-person state and task framing.

On top of context-specific fine-tuning, sampling protocols must avoid in-prompt self-conditioning and guard against order sensitivity. \textcite{ArgyleBusbyFuldaGublerRyttingWingate2023}’s one-subject-per-backstory template already exemplifies this sort of independence, while \textcite{park2024generativeagentsimulations1000}’s pipelines query anchored agents without recycling earlier synthetic outputs. In panel evaluations where LLMs or humans judge model answers, simply swapping left–right placement or list order can change which answer is preferred even when the content is identical, so a fixed order can create the appearance of convergence that is really a position effect \parencite{chen2024humansllmsjudgestudy, shi2025judgingjudgessystematicstudy}. Council-style studies quantify this risk by reporting bootstrap confidence intervals for rankings and win rates, which test whether the apparent winner persists when votes or items are resampled \parencite{li2024crowdsourceddatahighqualitybenchmarks}. Combined with evidence of position, length, and self-preference biases in LLM-as-a-judge settings, this motivates concrete safeguards in agent pipelines \parencite{wataoka2025selfpreference, zheng2023judgingllmasajudgemtbenchchatbot}. The implication is that if LLM agents are treated as predictors under explicit conditioning rather than as calibrated posteriors, independence checks, randomized item order, and fixed decoding settings should also be enforced so that any apparent agreement reflects the model’s predictive signal.

Stress-testing synthetic agents should be routine and openly disclosed, and researchers should recognize agents’ extrapolative limits. An LLM agent that performs well on in-domain tasks like matching known survey results may degrade when confronted with different novel stimuli, or even the same task worded differently. \textcite{park2024generativeagentsimulations1000}’s interview-anchored agents achieved high accuracy on reproducing GSS results, but broader evaluations suggest that such agents can falter outside the narrow domain they were tuned for. For example, \textcite{yeykelis2025usinglargelanguagemodels} conducted one of the largest cross-task replication studies to date, simulating over 19,000 AI participants based in Claude 3.5 across 133 behavioral effects. They found the models successfully reproduced about 76\% of known main effects, but only 27\% of interaction effects, highlighting a steep drop in reliability for more complex or out-of-distribution patterns . Likewise, \textcite{chen2025predictingfieldexperimentslarge} tested GPT-4/Claude on 319 real-world field experiments and noted strong overall accuracy but systematic failures on studies involving gender, culture, or identity-sensitive interventions. These findings emphasize that subgroup or context reliability varies by task and stimulus. \textcite{aher2023usinglargelanguagemodels} Turing Experiment framework probes this by introducing novel stimuli alongside classic experiments. While GPT-based agents accurately replicated famous results such as the Milgram obedience effect, a twist like the wisdom of crowds estimation task unveiled the aforementioned hyper-accuracy distortion, showing another clear divergence from expected human behavior. These results consistently prompt one of the most fundamental critiques of using LLMs as social agents—are the LLMs regurgitating material learned in pre-training, tuning, or prompting instead of modeling people? Does that even matter? It matters when the claim is about generalization, subgroup reliability, causal interpretation, or posterior beliefs about humans. A system that memorizes main effects can still fail on things like extrapolated interactions in shifted contexts, or on identity-sensitive interventions, and it can give falsely precise intervals around a biased mean. It matters less when the stated goal is narrow, in-domain prediction under explicit conditioning where overlap with past material is acceptable if disclosed and if out-of-domain checks are passed.

Researchers can treat this as a testing and reporting problem. Stress-tests such as varying the prompt phrasing, context, or experimental setup should be a standard part of validation. Publications may wish to include an out-of-distribution stress grid and report where the predictive mimicry breaks down. In addition, research would benefit from authors compiling a brief failure anthology describing cases where the synthetic agents produce implausible or biased outputs under stress. By disclosing these failure modes, researchers make clear the boundaries of their LLM agents’ validity. Robust stress-testing guards against overconfidence in synthetic data and helps readers understand exactly when and where LLM-driven simulations can be trusted and where they cannot.

Finally, rigorous reproducibility and full disclosure are not optional when using synthetic agent data. To build credibility, researchers should release all key details of their LLM simulation pipeline from prompt templates and persona/interview scripts, model identifiers and versions, decoding parameters like temperature and max tokens, and any random seeds, so that others can re-sample and recompute all reported quantities. It is equally important to demonstrate that results do not hinge on arbitrary choices like the ordering of questions or the particular random draw. For instance, independence and order-invariance checks such as shuffling prompt order and regenerating with new seeds can reveal lurking artifacts like position biases or self-consistency effects. If synthetic data from multiple runs or model instances are aggregated—which is becoming an increasingly common practice to stabilize outputs—authors should report any observed biases and explain how ensembling affected stability. Whenever an ensemble of LLM outputs is used as a sort of a council to improve reliability, the variance across members of that council and the degree to which averaging or voting alters the results should be documented.

\section{Conclusion}

This analysis is not intended to suggest that LLMs as synthetic agents have a limited future in social science. On the contrary, their promise is immense. As scaling continues to improve their capabilities, their power as near-term predictors of social phenomena becomes increasingly apparent. By effectively learning the complex patterns within vast datasets of human expression, these models can offer sophisticated quasi-interpolations of public attitudes and behaviors. Even more so, robust synthetic agents present an exciting frontier for research that is either ethically fraught or practically impossible to conduct in the real world, which could allow researchers to explore sensitive social dynamics or test counterfactual scenarios.
However, this promise can only be realized if the limitations are clearly recognized. The evidence that these models are not principled Bayesian reasoners and that their uncertainty remains an uninterpretable black-box must temper enthusiasm. The future of these tools is not the creation of a single, universal "LLM for social science," but the careful development of specific models tailored to specific social questions at the appropriate levels of analysis. The value lies in using them to guide policy and decision-making on near-term issues where a plausible interpolation of existing data is valuable. While the goal of one day achieving full mechanistic interpretability remains on the horizon, the current reality is that these models operate as black boxes whose internal logic is largely inaccessible. The responsible path forward, therefore, is to embrace these powerful new instruments not for what we hope they might one day become, but for the analytic tools they are now.

\printbibliography
\end{document}